\documentclass[letterpaper, 10 pt, conference]{ieeeconf}  

\IEEEoverridecommandlockouts                              

\usepackage{epsfig} 
\usepackage{mathptmx} 
\usepackage{times} 
\usepackage{amsmath} 
\usepackage{amssymb}  

\usepackage{color}
\usepackage{hyperref}
\usepackage{float}
\usepackage{graphicx}
\usepackage{algorithm}
\usepackage[noend]{algpseudocode}
\algblock{Input}{EndInput}
\algnotext{EndInput}
\algblock{Output}{EndOutput}
\algnotext{EndOutput}
\newcommand{\Desc}[2]{\State \makebox[3em][l]{#1}#2}

\title{\LARGE \bf
Learn and Link: Learning Critical Regions for Efficient Planning
}

\author{Daniel Molina, Kislay Kumar, Siddharth Srivastava\\
School of Computing, Informatics, and Decision Systems Engineering\\
Arizona State University\\
Tempe, Arizona 85281\\
}

\begin{document}

\maketitle
\thispagestyle{empty}
\pagestyle{empty}

\begin{abstract}

This paper presents a new approach to learning for motion planning (MP) where \emph{critical regions} of an environment are learned from a given set of motion plans and used to improve performance on new environments and problem instances. We introduce a new suite of sampling-based motion planners, \emph{Learn and Link}. Our planners leverage critical regions to overcome the limitations of uniform sampling, while still maintaining guarantees of correctness inherent to sampling-based algorithms. We also show that convolutional neural networks (CNNs) can be used to identify critical regions for motion planning problems. We evaluate Learn and Link against planners from the Open Motion Planning Library (OMPL) using an extensive suite of experiments on challenging motion planning problems. We show that our approach requires far less planning time than existing sampling-based planners.

\end{abstract}

\section{Introduction}
\begin{figure}[t!]
\centering
\includegraphics[scale=0.2]{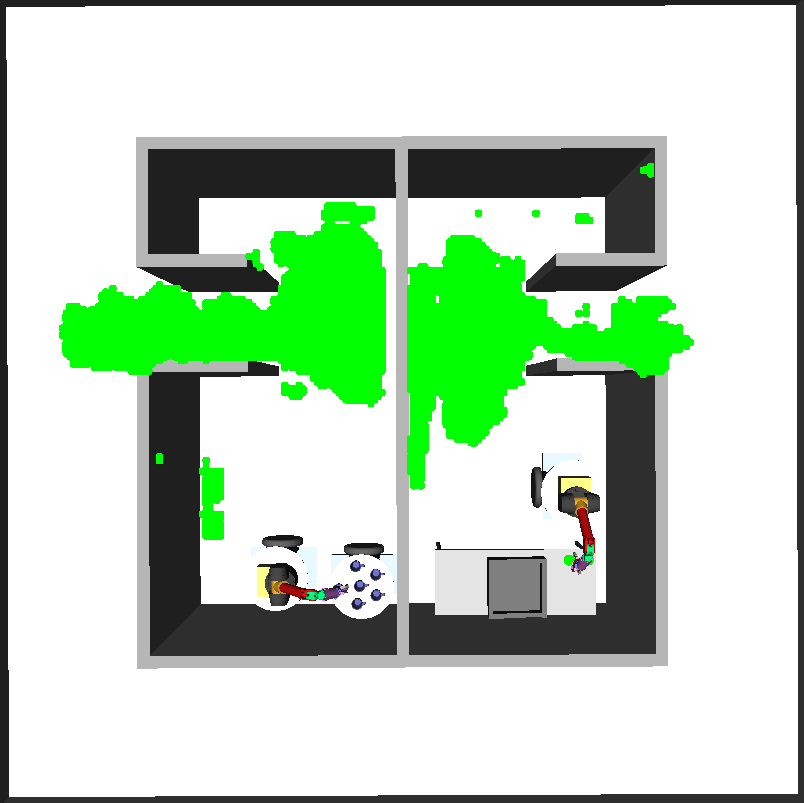}
\caption{Critical regions [green] predicted by our CNN for a transportation task through narrow channels.}
\label{llprmexample}
\end{figure}

The motion planning (MP) problem deals with finding a feasible trajectory that takes a robot from a start configuration to a goal configuration without colliding with obstacles. From a computational complexity point of view, even a simple form of the MP problem is NP-hard \cite{reif1979complexity}. In order to achieve computational efficiency, motion planning methods relax requirements of completeness. Sampling-based motion planners, such as Rapidly-exploring Random Trees (RRT) \cite{lavalle2001randomized} and Probabilistic Roadmaps (PRM) \cite{svestka1996probabilistic}, rely on \emph{probabilistic completeness}, which assures a solution, if one exists, as the number of samples approaches infinity. Sampling-based motion planners sample a set of states from the configuration space (C-space) and check their connectivity without ever explicitly constructing any obstacles. This can reduce computation time considerably, especially as environments increase in complexity. Their performance, however, hinges on the distribution from which points in the C-space are sampled. Uniform samplers can fail in common situations, such as in Figure \ref{llprmexample}, where the robot needs to traverse narrow regions of measure close to zero under a uniform density in the C-space. \par

In this work, we propose a new version of sampling-based motion planners with associated learning paradigms that inherit the probabilistic completeness properties of RRTs and PRMs, and are designed to be able to utilize learned sampling distributions. In particular, our Learn and Link (LL) suite of planners can utilize learned information about \emph{critical regions} of the C-space, which are less likely to be sampled under a uniform distribution (e.g., narrow corridors \cite{lindemann2005current}) but are critical since most solutions for a given, desired class of problems pass through them. This relates to the notion of \emph{landmarks}, or parts of the state space that are necessary for reaching the goal in discrete planning problems \cite{hoffmann2004ordered}. However, critical regions are not only useful (albeit not necessary) for reaching the goal, but are also less likely to be reached under a stochastic search paradigm. \par

\begin{figure}[thpb]
\centering
\includegraphics[scale=0.3]{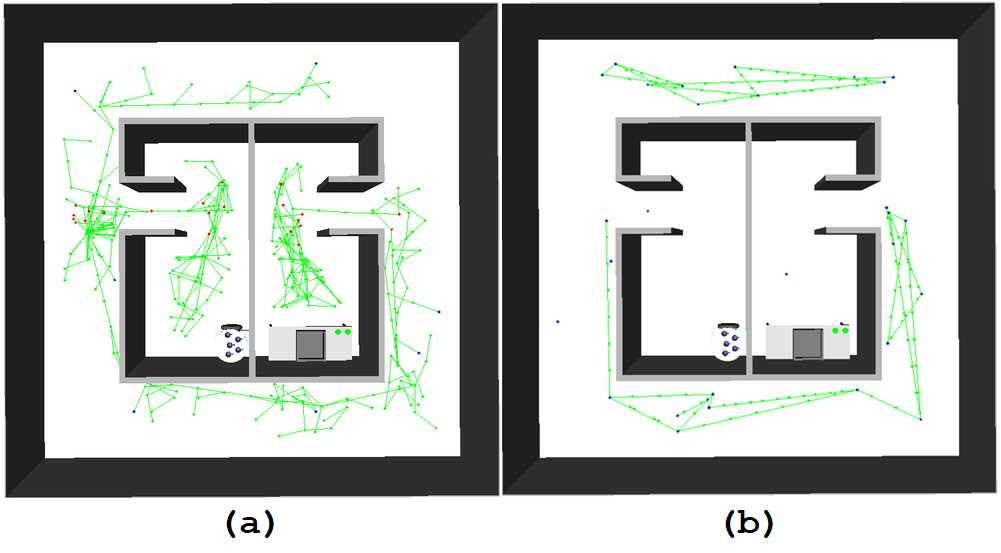}
\caption{An example of a roadmap created using LL-RM (a) versus a vanilla PRM (b) for a Barrett WAM arm transportation task. The green points are states that were created when linking the vertices of the roadmap, the blue points are the vertices of the roadmap that were uniformly sampled, and the red points are the vertices of the roadmap that came from the critical regions. This illustrates the extreme difference in C-space coverage using the same amount of time.}
\label{roadmapexample}
\end{figure}

Naive approaches for using critical regions (either learned or hand-coded) in existing sampling-based motion planners tend to fail: merely increasing the probability of sampling from such regions does not improve the performance of RRT planners because the parts of the tree(s) closest to the critical regions tend to be those that are crashing into the walls adjacent to them. Even when allowing a PRM planner to select configurations from critical regions as vertices in its roadmap, its simple local planner is unable to connect the vertices in critical regions to those uniformly sampled unless a considerable amount of time is used in the roadmap building process (see Figure \ref{roadmapexample}). The LL planners presented in this paper leverage the positives of these planners while having the necessary modifications to properly utilize critical regions. \par  

We also present a new approach for learning critical regions using convolutional neural networks (CNNs)~\cite{badrinarayanan2015segnet,ronneberger2015u}. We show that when used with our LL planners, this approach can lead to immense speedups in motion planning when image-based training data is available for the planning environment. Since our model only gives base poses, when dealing with higher dimensional planning we append each configuration pulled from the critical regions with a random, collision-free configuration for the additional DOF values prior to calling our planners. Although this learning pipeline is limited to image-based representations, our planners use critical regions as inputs and can work with any approach that provides an estimate of the critical regions for a given environment. Our approach is advantageous over pure sampling-based planners and pure learners: it leverages learning from experience to outperform sampling-based planners, but avoids the possibility of missing solutions that limits pure imitation learning, and remains probabilistically complete. \par
 
The rest of this paper is organized as follows. We begin with a survey of prior work that our approach draws upon in Section II. Section III defines the notion of critical regions; followed by Section IV, which presents the LL planners and discusses their properties. Section V presents our approach for learning critical regions using CNNs. Finally, Section VI presents the results of our empirical evaluation on a range of motion planning problems. \par

\section{Related Works}
Several methods have been proposed for guiding sampling-based motion planners to solutions. Heuristically-guided RRT \cite{urmson2003approaches} uses a probabilistic implementation of heuristic search concepts to create a reasonable bias towards exploration, as well as exploiting known good paths. Although this approach was able to produce less expensive paths, it incurs a high computational cost. Anytime RRTs \cite{ferguson2006anytime} reuse information from previous RRTs to improve on the path by rejecting samples that have a higher heuristic cost. Batch Informed Trees (BIT*) \cite{gammell2015batch} use a heuristic to efficiently search in a series of increasingly dense implicit random geometric graphs while reusing previous information. In contrast, our method guides our sampling-based motion planners to solutions without the need of a heuristic by leveraging learned critical regions. \par

The coupling of learning and MP has been extensively investigated in the past. As discussed in the introduction, naive approaches for using learning to bias sampling in stochastic motion planners don't perform well. Recent work by Ichter et al. uses a conditional variational autoencoder to bias sample points for MP conditioned on encoded environment variables \cite{ichter2018learning}. This encoding is generalizable to higher dimensions. However, it requires structuring the data to encompass the state of the robot, the environment, the obstacles (encoded as an occupancy grid), and the start and goal configurations. Moreover, during inference, the network model requires this expensive data structuring again, which can take around 50 seconds. In contrast, we focus on image-based learning where data can be easily generated for training using a top-view camera. Moreover, inferences can also be made using a top-view image of the environment in less than 5 seconds. Havoutis et al. use topology to learn sub-manifold approximations that are defined by a set of possible trajectories in the C-space \cite{havoutis2009motion}. This requires either motion plans that are generated through a motion capture device, or hand-crafted partial plans. Pan et al. use instance-based learning where prior collision results are stored as an approximate representation of the collision space and the free C-space \cite{pan2013faster}. This is used to make cheaper probabilistic queries. Although their method shows significant improvement in some environments, their work is limited in finding solutions through narrow passages between obstacles where the optimal solution may lie. In our work, the network learns the positions of regions that are critical for a given class of MP problems, but have a low probability of getting sampled under a uniform distribution, such as narrow regions. The closest related work to this paper presented a preliminary version of our methods in a non-archival venue \cite{molina2019planrob}. The current paper rigorously develops and presents our full approach. \par

\section{Formal Framework}
Given a robot \emph{R}, an environment \emph{E}, and a class of MP problems \emph{M}, we define the \emph{measure of criticality} of a Lebesgue-measurable open set $r\subseteq \mathbb{R}^n$, $\mu(r)$, as  $lt_{s_n \rightarrow ^+ r}\frac{f(r)}{v(s_n)}$, where \emph{$f(s_n)$} is the fraction of observed motion plans solving tasks from \emph{M} that pass through $s_n$, $v(s_n)$ is the measure of $s_n$ under a reference (usually uniform) density, and $\rightarrow^+$ denotes the limit from above along any sequence $\{s_n\}$ of sets containing $r$ ($r\subseteq s_n$ for all $n$). Note that $\mu(r)$ is zero when $f(r)=0$. While $\mu(r)$ can be infinite for a region, for all practical purposes we consider regions $r$ with $v(r)>0$ under the uniform density. Intuitively, regions with high criticality measures are those that are vital for solutions to problems in \emph{M}, but have a low probability of being sampled under a uniform density. \par

\section{Learn and Link Planners}
In this section we discuss the methods that make up our planners. We describe both the single query planner, LLP, and the multi-query planner, LL-RM. \par

\subsection{Learn and Link Planner}
LLP is Learn and Link's single query planner. This version differs from LL-RM in that we do not seed our roadmap using vertices that were uniformly sampled (i.e. $m=0$), and we pass in the start and goal configurations right away to algorithm \ref{build}. We do this so that instead of building a general roadmap that spreads across the entire environment, we build a biased roadmap in which subgraphs rooted from the start and goal configurations are connected using additional subgraphs rooted at critical regions to speed up single query planning. \par

We first describe algorithm \ref{build} in LLP mode. In lines $14-17$, $n$ random collision-free configurations are added as vertices to the roadmap from the critical regions identified by the model. Since we are in LLP mode, in lines $18-21$, 0 configurations are added as vertices to the roadmap using a uniform sampler. In lines $22-26$, subgraphs rooted from the start and goal configurations are added to the roadmap. For the remainder of the algorithm, we attempt to link the subgraphs spawned from the vertices in the roadmap. In line $28$, a random sample is taken to grow the current subgraph in its direction. In line $29$, an attempt is made to extend the current subgraph to $q_{new}$, a new configuration in the direction of $q_{rand}$. If adding $q_{new}$ to the graph results in a collision, $i.e.$ \emph{EXTEND} returns $Trapped$, $q_{new}$ is not added to the graph; otherwise it is added. In line $30$, a connectivity attempt occurs to link the current subgraph to the remaining graphs in the roadmap; once all the subgraphs have been connected, \emph{Linked} is returned and the roadmap is complete. By this point, since we are in LLP mode, the start and goal configurations have been linked into a single graph. To extract a path \emph{P} connecting both points we use Dijkstra's algorithm \cite{dijkstra1959note} in line $32$. If the conditions in lines $29-30$ are not satisfied, we shift to the next subgraph in the roadmap, using a round-robin approach, in line $36$. If an explicit sample cap is reached, $i.e.$ $S \neq \infty$, without a solution path being found, an empty path, indicating a failure, is returned. \par

\begin{algorithm} [t]
\caption{\small \texttt{Learn and Link}}\label{build}
\begin{algorithmic}[1]
\begin{footnotesize}
\Input
  \Desc{$N$:}{number of critical region states to include}
  \Desc{$M$:}{number of uniform states to include}
  \Desc{$CR$:}{list of critical region points}
  \Desc{$Mode$:}{planner mode; LLP or LL-RM}
  \Desc{$Q_{start}$:}{start configuration, if LLP mode}
  \Desc{$Q_{goal}$:}{goal configuration, if LLP mode}
\EndInput
\Output
  \Desc{$P$:}{collision-free path from $q_{start}$ to $q_{goal}$, if it exists}
  \Desc{$RM$:}{constructed roadmap}
\EndOutput
\Procedure{\texttt{LL(n,m,CR,mode,q$_{start}$,q$_{goal}$)}}{}
    \State $curr \gets 0$
    \State $RM \gets []$
    \For {$n = 0 \textbf{ to } N-1$}
        \State $s \gets SAMPLE(CR)$
        \State $G_{n}.init(s)$
        \State $RM.append(G_{n})$
    \EndFor
    \For {$m = 0 \textbf{ to } M-1$}
        \State $s \gets SAMPLE()$
        \State $G_{N+m}.init(s)$
        \State $RM.append(G_{N+m})$
    \EndFor
    \If {$mode==LLP$}
        \State $G_{N+M}.init(q_{start})$
        \State $G_{N+M+1}.init(q_{goal})$
        \State $RM.append(G_{N+M})$
        \State $RM.append(G_{N+M+1})$
    \EndIf
    \For {$s = 1 \textbf{ to } S$}
        \State $q_{rand} \gets UNIFORM()$
        \If {$EXTEND(G_{curr},q_{rand}) \neq Trapped$}
            \If {$LINK(RM,G_{curr},q_{new})==Linked$}
                \If{$mode==LLP$}
                    \State $P \gets PATH(RM[0])$
                    \State {\bf Return} \textit{P}
                \Else
                    \State {\bf Return} \textit{RM}
                \EndIf
            \EndIf
        \EndIf
        \State $G_{curr} \gets SWAP(RM,G_{curr)}$
    \EndFor
    \State {\bf Return} []
\EndProcedure
\end{footnotesize}
\end{algorithmic}
\end{algorithm}

Algorithm \ref{link} is used in an attempt to link a subgraph to the remaining graphs in the roadmap (lines $9-11$), to remove dead graphs from consideration (line $12$), and to check whether all the subgraphs in the roadmap have been linked (line $13$). A subgraph is considered dead once it has been linked and added to another graph. Once only one graph remains in the roadmap list, \emph{Linked} is returned to indicate that the roadmap is connected. \par

\begin{algorithm} [t]
\caption{\small \texttt{LINK}}\label{link}
\begin{algorithmic}[1]
\begin{footnotesize}
\Input
  \Desc{$RM$:}{roadmap of graphs to be connected}
  \Desc{$G_{curr}$:}{current subgraph being grown}
  \Desc{$Q_{new}$:}{most recent configuration added to G$_{curr}$}
\EndInput
\Output
  \Desc{$S$:}{status of $G_{curr}$'s link attempt}
\EndOutput
\Procedure{\texttt{LINK(RM,G$_{curr}$,q$_{new}$})}{}
    \State $R \gets []$
    \For {$G_{i} \textbf{ in } RM \setminus G_{curr}$}
        \If{$CONNECT(G_{i},q_{new})==Reached$}
            \State $R.append(G_{i})$
        \EndIf
    \EndFor
    \State $RM.link\char`_and\char`_remove(R,G_{curr})$
    \If{$|RM| == 1$}
        \State $S \gets Linked$
    \ElsIf{$|R| > 0$}
        \State $S \gets Connected$
    \Else
        \State $S \gets Advanced$
    \EndIf
    \State {\bf Return} \textit{S}
\EndProcedure
\end{footnotesize}
\end{algorithmic}
\end{algorithm}

The \emph{CONNECT} and \emph{EXTEND} algorithms are reused and adapted from RRT-Connect to work with graphs instead of trees. They are used to grow the current subgraph in the direction of the random samples taken. \par



\subsection{Learn and Link Roadmap}
LL-RM is Learn and Link's multi-query planner. This version differs from LLP in that we attempt to build a general roadmap which can be reused multiple times for traversing a C-space based on collision-free configurations from critical regions, as well as some uniformly sampled. To solve a query, we simply try to connect the start and goal configurations to the roadmap given by algorithm \ref{build} in LL-RM mode. If we are successful, we use Dijkstra's algorithm to obtain a plan. \par

When in LL-RM mode, the linking process works the same as LLP. The only differences in LL-RM is that we include additional vertices in the roadmap from areas which were uniformly sampled (i.e. $m>0$) in lines $18-21$, and we return the roadmap \emph{RM}, instead of a path, when the subgraphs are connected in line $35$. \par

Algorithm \ref{plan} is the planning component of LL-RM. In lines $9-12$, two subgraphs are initialized from the start ($q_{start}$) and goal ($q_{goal}$) configurations in an attempt to connect them to the existing roadmap \emph{RM}. In lines $13-18$, the same approach used in the building process is employed to connect the start and goal subgraphs to the roadmap. In line $16$, a solution check occurs. If a solution is found, the path \emph{P} connecting the start and goal configurations is obtained using Dijkstra's algorithm in line $17$. \par 

\begin{algorithm} [t]
\caption{\small \texttt{LL-RM PLAN}}\label{plan}
\begin{algorithmic}[1]
\begin{footnotesize}
\Input
  \Desc{$Q_{start}$:}{start configuration}
  \Desc{$Q_{goal}$:}{goal configuration}
  \Desc{$RM$:}{roadmap created using LL in LL-RM mode}
\EndInput
\Output
  \Desc{$P$:}{collision-free path from $q_{start}$ to $q_{goal}$, if it exists}
\EndOutput
\Procedure{\texttt{RM-PLAN(q$_{start}$,q$_{goal}$,RM)}}{}
    \State $curr \gets 0$
    \State $G_{1}.init(q_{start})$
    \State $G_{2}.init(q_{goal})$
    \State $RM.append(G_{1})$
    \State $RM.append(G_{2})$
    \For {$s = 1 \textbf{ to } S$}
        \State $q_{rand} \gets UNIFORM()$
        \If {$EXTEND(G_{curr},q_{rand}) \neq Trapped$}
            \If {$LINK(RM,G_{curr},q_{new})==Linked$}
                \State $P \gets PATH(RM[0])$
                \State {\bf Return} \textit{P}
            \EndIf
        \EndIf
        \State $G_{curr} \gets SWAP(RM,G_{curr)}$
    \EndFor
    \State {\bf Return} []
\EndProcedure
\end{footnotesize}
\end{algorithmic}
\end{algorithm}

\subsection{Probabilistic Completeness}
The LL planners maintain the probabilistic completeness property inherent to sampling-based motion planners. Since LLP and LL-RM only add a finite set of points to seed their roadmaps, the added critical region vertices do not reduce the set of support (regions with non-zero probability) of its uniform sampler, and thus, this property is preserved. \par

\subsection{Distinction Between LL Planners and PRM}
LLP and LL-RM work analogously to PRM, but they have their differences. Essentially, PRM views each initial random configuration as a vertex of a larger graph, whereas LLP and LL-RM view each initial configuration as the start of its own graph. Also, PRM uses a quick and simple local planner to connect its vertices, so they do not store their local plans; whereas LLP and LL-RM use a tree-based local planner inspired by RRT-Connect which saves all the spawned branches when connecting two vertices and incorporates them into the roadmap. Seeing as we store the trees created by the local planner, instead of simply attempting to connect a vertex to others within a neighborhood like in PRM, we leverage the additional connectivity and attempt to connect each subgraph to every other non-connected subgraph. These differences are necessary so that subgraphs spawned from critical regions can be linked to those outside these regions. We discovered that since critical regions are in areas that are unlikely to be reached under a stochastic search paradigm, it was not enough to use a simple local planner to link the two types of subgraphs; but once they were linked, they drastically sped up the planning process. \par

\section{Learning Critical Regions}
To learn critical regions, we use an image-based approach. This consists of two phases: a data generation phase and a model training phase. \par

\begin{figure}[thpb]
\centering
\includegraphics[scale=0.275]{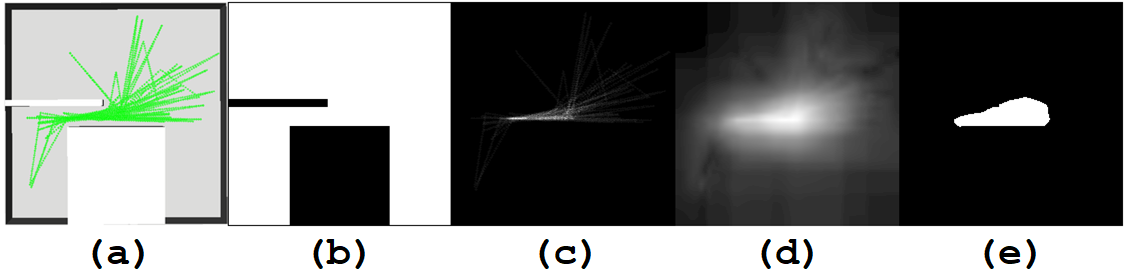}
\caption{(a) An example training environment overlain with motion traces. (b) Model input obtained post raster scan. (c) Motion trace image based on $\mu$-criticality of each pixel. (d) Saliency map obtained from the motion trace image. (e) Label obtained after binning the saliency map based on pixel intensity.}
\label{inputex}
\end{figure}

\subsection{Data Generation}
For each instance of an environment, we begin by randomly selecting a set of 50 motion planning problems from \emph{M} $\{\Pi_{1},...,\Pi_{50}\}$ and running an off-the-shelf motion planner to generate a corresponding set of motion plans $\{\tau_{1},...,\tau_{50}\}$. We repeat this process multiple times for each handmade environment to make sure we fully cover its critical regions; 179 times per environment in our dataset. In our data generation process, we utilize an OpenRAVE \cite{diankov2008openrave} implementation of OMPL's RRT-Connect planner by \href{https://github.com/personalrobotics/or_ompl}{\texttt{https://github.com/personalrobotics}}, though any motion planner can be used instead. \par

We construct the 224x224 training images for each instance using a raster scan and a saliency model. We describe the process for an \emph{SE(2)} robot (see Figure \ref{inputex}), though it can be extended to mobile manipulators, such as the Barrett arm on a mobile base. We begin by creating a pixel-sized obstacle based on the dimensions of the desired image and the bounds of a given environment. We proceed by scanning the pixel-sized obstacle across the environment. For the input images, if a collision is detected with an environment's obstacles, we select a black pixel, otherwise a white pixel is selected (see Figure \ref{inputex}b). For the motion trace images, we assign a pixel value based on the $\mu$-criticality of the region the pixel encompasses, which we obtain using $\{\tau_{1},...,\tau_{50}\}$ (see Figure \ref{inputex}c). We then use an implementation of Itti's saliency model \cite{itti2000saliency} by \href{https://github.com/mayoyamasaki/saliency-map}{\texttt{https://github.com/mayoyamasaki}} to extract relevant salient information and smooth out the salient areas from the motion trace images (see Figure \ref{inputex}d). Finally, the saliency maps are binned into two categories, high saliency (denoted by white pixels) and low saliency (denoted by black pixels), and are used as the labels (see Figure \ref{inputex}e). \par


\begin{figure}[thpb]
\centering
\includegraphics[scale=0.15]{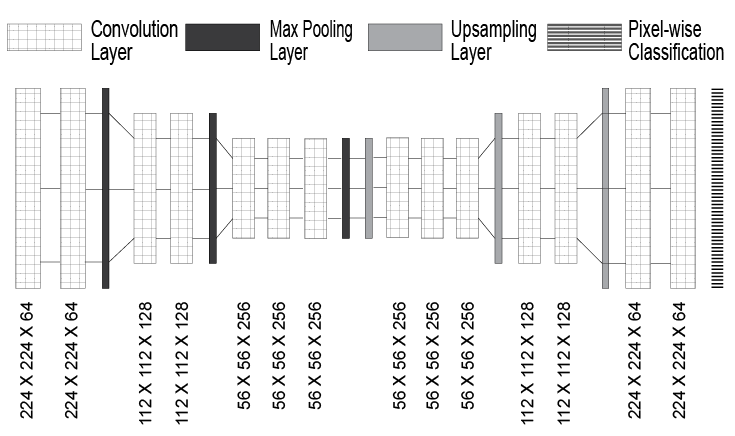}
\caption{Network architecture selected for our model.}
\label{arch}
\end{figure}

\subsection{Network Architecture}
We propose a general structure for a convolutional encoder-decoder neural network which learns to detect critical regions. \par

Our network, depicted in Figure \ref{arch}, has 14 convolutional layers, with 7 layers in the encoder network and 7 layers in the decoder network forming the encoder-decoder architecture for pixel-wise classification. A max pooling layer with stride 2 is introduced after each group of same number of filters to encode the learned representation. Similarly, an upsampling layer is added before each deconvolutional layer group of same number of filters. We draw inspiration from \cite{badrinarayanan2015segnet} for a learnable upsampling layer in the decoder network. \par

\begin{figure}[thbp]
\centering
\includegraphics[scale=0.45]{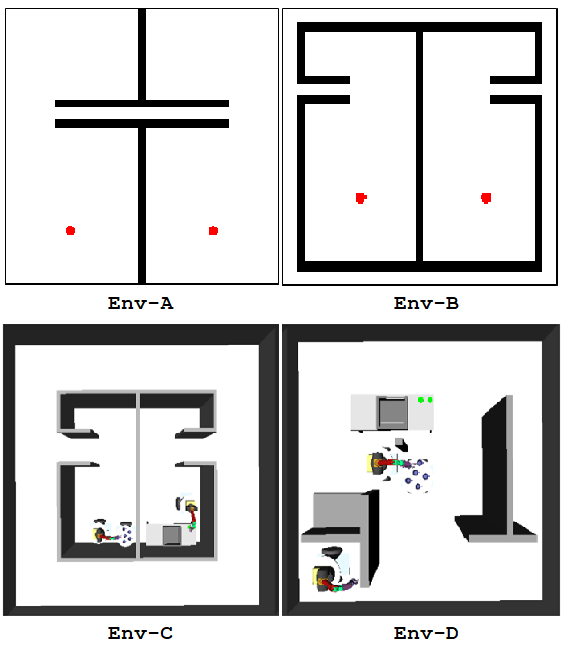}
\caption{Env-A and Env-B are the \emph{SE(2)} environments used to evaluate the model. Red dots represent the start and goal configurations. Env-C and Env-D are the 10-DOF test environments used to evaluate the model. The robot is placed at the start and goal configurations. Test environments are unseen to the network (they were not used in the training set of problems).}
\label{testex}
\end{figure}

\begin{figure}[thbp]
\centering
\includegraphics[scale=0.525]{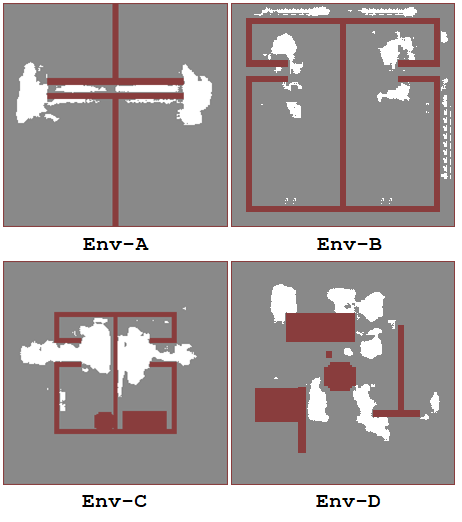}
\caption{Network output for the test environments shown in Figure 5. The $\mu$-criticality of the output is as follows: $\mu$(Env-A) = 0.604, $\mu$(Env-B) = 1.351, $\mu$(Env-C) = 0.538, and $\mu$(Env-D) = 0.398.}
\label{networkout}
\end{figure}

The first two convolutional layers have 64 filters with a $3 \times 3$ kernel. Motivated by recent promising results \cite{simonyan2014very}, we stack 3 layers with $3 \times 3$ kernel size to obtain a similar receptive field as a $7 \times 7$ kernel, with 81\% less parameters, and more effective training owing to the added non-linearity after every layer. For the initial layer group of filter size 64 and 128, we stack only two layers of kernel size $3 \times 3$. Though the receptive field is smaller than a $7 \times 7$ kernel, we still stack only 2 layers as our problem statement does not require learning extremely complex geometric features. The next 2 layers are of 128 filters with a $3 \times 3$ kernel. We add 3 layers of 256 filters each, with a $3 \times 3$ kernel, for a larger receptive field since deeper layers learn invariant complex features \cite{zeiler2014visualizing}. All the convolutional and max-pool layers have padding added to them. \par

\begin{figure*}[t!]
\centering
\includegraphics[scale=0.525]{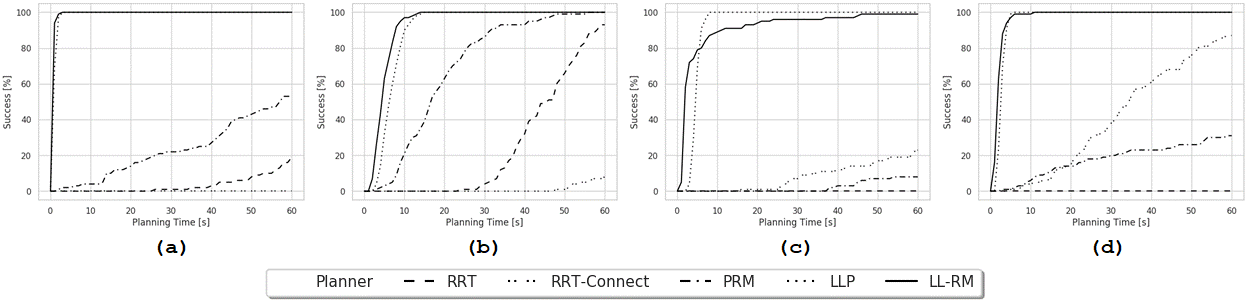}
\caption{Each plot shows show the percentage of trials solved versus planning time for the 100 trials for the corresponding environments in Figure \ref{testex}.}
\label{results}
\end{figure*}

In the decoder network, corresponding deconvolutional layers to the encoder network are used. The upsampled output is used for pixel-wise classification using a softmax cross-entropy loss function. Each layer in the network is activated using ReLu nonlinearity. \par 

Ablation analysis showed that shallower CNNs and deeper SegNet \cite{badrinarayanan2015segnet} inspired architectures predicted regions with lower criticality scores on average than our selected architecture \cite{llp_tech_report}. \par

\subsection{Training}
The network was trained using a mini-batch size of 16 and a dataset of 10,024 images created from 14 distinct environments with four rotations each. Following \cite{ioffe2015batch}, we did not train the network with dropout \cite{srivastava2014dropout} since the output of every layer is batch-normalised, which also acts as a regularizer. We use Adam Optimizer \cite{kingma2014adam} with a 0.1 learning rate to train the network. The network was trained for 50,000 epochs since the loss converges at this point. The training images are shuffled before each epoch and trained with mini-batch to ensure that every input to the network is different from the previous. This assists the optimizer to exit local minima. We used an implementation of SegNet by \href{https://github.com/andreaazzini/segnet.tf}{\texttt{https://github.com/andreaazzini}} for its data pipelines since they provide a fast and efficient input pipeline which reduces training time. \par

On average, training for the full dataset takes approximately 3 hours on a single Nvidia GTX 1080Ti. \par

\section{Empirical Evaluation}
Our empirical analysis focuses on investigating two main questions:
\begin{enumerate}
    \item Can CNNs be used to identify critical regions for motion planning?
    \item Can critical regions be used to improve planning performance?
\end{enumerate} \par

The first consideration aims to see if we can extend the visual prowess exhibited by CNNs to identifying the critical regions of an environment. The second consideration aims to see if knowing critical regions helps a planner reduce its computation time. Our intent is not to create the best, optimal planner, but to evaluate the gains that can be made when a planner properly leverages the critical regions of the C-space being traversed. \par

To investigate these considerations, we designed challenging MP problems for \emph{SE(2)} and the Barrett WAM arm (see Figure \ref{testex}), and we explored various network architectures. We evaluate the quality of the critical regions given by our CNN using their measure of criticality $\mu$ (see Figure \ref{networkout}), as well as the planning time used by our planners. 100 MP problems were constructed for evaluation using the same start and goal pair, the same range, and a planning time limit of 60 seconds. LLP and LL-RM both use 5\% of the critical regions identified as $n$, and \emph{m} is $0$ and $n/10$, respectively. LL-RM and OMPL's PRM are both given 1 second to build a roadmap prior to planning. Our approach is for robots with omnidirectional base movements, though any movement constraints can be added in the \emph{EXTEND} module. It is important to note that OMPL is written in highly optimized C++ code compared to our Python implementation. \par

Our results suggest that both LLP and LL-RM require far less time to obtain a solution than OMPL's RRT, RRT-Connect, and PRM planners. Figure \ref{results} shows a comparison of planning time used by the OMPL planners and our LL planners. The complete source code for our planners is available at \href{https://aair-lab.github.io/ll.html}{\texttt{https://aair-lab.github.io/ll.html}}. \par

\subsection{SE(2) Domain}
For \emph{SE(2)}, LLP and LL-RM outperformed OMPL's planners in terms of average planning time and success rate. On Env-A, RRT, RRT-Connect, and PRM had success rates of 19\%, 0\%, and 53\%, respectively; whereas the LL planners were 100\% successful. For successful plans, LLP and LL-RM required 97\% and 99\% less time on average, respectively, than PRM, the best performing OMPL planner on this environment. On Env-B, RRT, RRT-Connect, and PRM had success rates of 93\%, 8\%, and 100\%, respectively; whereas the LL planners were 100\% successful. For successful plans, LLP and LL-RM required 64\% and 74\% less time on average, respectively, than PRM, the best performing OMPL planner on this environment. The low success rates of the RRT planners are a result of their trees' inability to cross narrow channels because nodes adjacent to samples in the channels tend to be on the opposite side of the wall. \par

\subsection{10-DOF Domain}
For the transportation tasks using the movable Barrett arm, LLP and LL-RM require less planning time on average and had higher success rates than the OMPL planners. On Env-C, RRT, RRT-Connect, and PRM had success rates of 0\%, 23\%, and 8\%, respectively; whereas LLP and LL-RM had success rates of 100\% and 99\%, respectively. When comparing successful plans, both LLP and LL-RM required 88\% less time on average than RRT-Connect, the best performing OMPL planner on this environment. On Env-D, RRT, RRT-Connect, and PRM had success rates of 0\%, 87\%, and 31\%, respectively; whereas the LL planners were 100\% successful. When comparing successful plans, LLP and LL-RM required 89\% and 92\% less time on average, respectively, than PRM, the best performing OMPL planner on this environment. As in \emph{SE(2)}, similar issues hindered the RRT planners' performance in the 10-DOF domain. \par

\section{Conclusions}
We presented a new approach to learning for MP and used it to create a new suite of sampling-based motion planners, Learn and Link. We also constructed a fully convolutional encoder-decoder neural network to learn critical regions for MP problems that generalizes across different domains. Our model is used by our LL planners to remedy the limitations of uniform sampling without compromising guarantees of correctness. \par


\section*{ACKNOWLEDGMENT}
This work was supported in part by the NSF under grant IIS 1909370. \par

\clearpage
\bibliographystyle{IEEEtran} 
\bibliography{ref}

\end{document}